\documentclass[11pt]{article}

\usepackage[preprint]{acl}

\usepackage{times}
\usepackage{latexsym}

\usepackage[T1]{fontenc}

\usepackage[utf8]{inputenc}

\usepackage{microtype}

\usepackage{inconsolata}

\usepackage{graphicx}
\usepackage{booktabs}
\usepackage{amsmath}
\usepackage{subcaption}
\usepackage{amsfonts}
\usepackage{enumitem}
\title{When Models Refuse: Political Steerability and Feature Richness as Measures of Ideological Depth}

\author{Shariar Kabir\\ [5pt]
Bangladesh University of Engineering and Technology\\ [5pt]
\texttt{shariar1405076@gmail.com}
}

\begin{document}
\maketitle
\begin{abstract}
Large language models (LLMs) sometimes refuse to follow benign instructions, such as declining to argue a political position or adopt a stated persona, and such refusals are commonly read as safety guardrails at work. We ask whether they can instead signal a \textbf{capability deficit}: a shortage of the internal representations a model needs to reason from the instructed perspective. To investigate, we introduce \textbf{ideological depth}, a property with two components: (i) a model's ability to follow political instructions without \textbf{failure} (steerability), and (ii) the \textbf{feature richness} of its internal political representations, measured with sparse autoencoders (SAEs). Using two widely used openweight LLMs as candidates, we compare interventions based on prompts and activation-steering, and probe political features with publicly available SAEs. We find large, systematic differences: a model that is more steerable in both ideological directions activates \textbf{$\sim 7.3\times$} more distinct political features, while the other model instead responds with increased refusals. Causally ablating a small, targeted set of political features from the former model reproduces the same feature-poor behavior and drives up refusals. Together, these results indicate that refusals on benign prompts can arise from \textbf{capability deficits} rather than fixed safety rules, and that ideological depth is a measurable property of LLMs that helps predict when a model will refuse.

\end{abstract}

\section{Introduction}
Large Language Models (LLMs) are increasingly used to summarize, argue, and advise on political topics, making their socio-political behavior a subject of critical importance. A growing body of research confirms that LLMs exhibit discernible political ideologies and encode interpretable ideological representations within their intermediate layers \cite{santurkar2023whose, kim2025linearrepresentationspoliticalperspective}, attracting attention from socio-political researchers who are beginning to leverage these models as tools for analysis \cite{linegar2023large, aldahoul2025large}. Yet these models behave inconsistently: when asked to argue a position or adopt a political persona, a model will sometimes comply and sometimes \textit{refuse}, even on entirely benign prompts. Such refusals are routinely attributed to safety alignment, but it is rarely asked whether they instead reflect something the model \textit{cannot} do rather than something it is trained not to do.
Prior work has shown that an LLM's political output can be significantly shifted through simple prompt engineering, such as applying argumentative pressure \cite{kabir2025wordsreflectbeliefsevaluating}, but the degree of this malleability varies sharply across models and topics. Attempts to steer a model toward a viewpoint can have divergent outcomes: where one model adjusts its response to reflect the instructed viewpoint, another defaults to a higher rate of refusal, especially on contentious topics.

This raises a crucial question: \textit{when a model refuses a benign instruction, is it exercising a safety guardrail or revealing the limits of its internal representations?}
We address this by measuring a model's \textbf{ideological depth} using political instructions as a test bed, a property capturing both (1) how reliably a model can adopt instructed political personas without refusing, and (2) the breadth and coherence of political features present in its activations. Our central hypothesis is that deeper political structure yields (a) higher-quality compliance with conflicting ideological instructions and (b) stable, causal features whose manipulation produces predictable changes in behavior—and, conversely, that a model lacking such structure falls back on refusal.

We evaluate two open models: Llama-3.1-8B-Instruct and Gemma-2-9B-IT using a two-pronged approach: (i) steerability via prompt engineering and activation addition, and (ii) feature richness via SAEs trained on the target layers reported to hold political features. We test across 12 topics using 126 evaluation prompts, each presented in \textit{three variants}: original, with a supporting argument, and with a counter-argument.
Our findings are consistent across methods. Gemma exhibits broader bidirectional steerability, while Llama often responds to conservative steering with refusals. Our SAE analyses reveal that Gemma activates far more distinct, thematically coherent political features. Furthermore, targeted ablations of the salient features increase the refusal rates across topics, supporting the view that refusals can stem from insufficient representations, not only from safety policy.

Our contributions are: (1) A definition and measurement protocol for ideological depth combining steerability and SAE-based feature richness. (2) Evidence that depth differs substantially across similarly sized models and predicts refusal behavior under political instructions. (3) Causal tests showing that ablating a compact set of political features increases refusal, directly supporting our hypothesis.

\section{Related Works}


The existence of political biases in LLMs is well-documented. Early work demonstrated that models trained on large internet corpora inevitably absorb and reflect the societal biases present in that data \cite{bender2021dangers, caliskan2017semantics}. More recent studies have specifically quantified the political leanings of various models, often by having them complete standard political science questionnaires, revealing consistent ideological alignments \cite{santurkar2023whose, peng2025partisanleaningcomparativeanalysis}. Findings show that these ideologies are not merely surface-level phenomena but are encoded as features in the models' internal activations. \citet{gurnee2023language}, for example, used linear probes to uncover learned representations of geographical and historical concepts in activation space. Similarly, \citet{kim2025linearrepresentationspoliticalperspective} uses linear probes revealing representations for political perspectives in LLMs. Our contribution is to move beyond simply identifying these features and to instead analyze the robustness and complexity of the structures they form.

A significant amount of work has explored methods to control LLM outputs. Prompt engineering is the most straightforward approach, where the user's input is crafted to guide the model toward a desired response style or content. However, its limitations in achieving consistent control motivate more direct intervention methods \cite{nanda2023attribution}. Activation steering, or activation addition, has emerged as a powerful technique for modifying model behavior at inference time without retraining. By adding a "steering vector" -- generally derived from the difference in activations between contrasting concepts (e.g., "love" vs. "hate") -- to the residual stream, researchers have been able to steer models towards specific emotions, topics, or styles \cite{rimsky-etal-2024-steering, zou2025representationengineeringtopdownapproach}. 

Our work utilizes these established steering techniques not as an end in themselves, but as a diagnostic tool to measure a model's ideological resilience, which we then correlate with its internal structure.
To analyze the internal structure of LLMs, we draw heavily from the field of mechanistic interpretability. A central challenge in this field is that models represent more features than they have neurons, a phenomenon known as superposition \cite{elhage2022toymodelssuperposition}. Sparse Autoencoders (SAEs) have recently become a leading method for untangling these representations by decomposing activations into an overcomplete basis of sparsely activating, monosemantic features \cite{bricken2023monosemanticity, templeton2024scaling}. Our work applies this cutting-edge technique to a new domain: identifying latent political features. We then perform causal interventions on these features through ablation to verify their effect on model behavior. While prior work has used linear probes to identify political features in a supervised manner \cite{kim2025linearrepresentationspoliticalperspective}, we use SAEs to find a more granular set of features and test their causal role in a model's ideological reasoning, allowing us to build a more detailed picture of its "ideological depth."

\section{Methodology}
\subsection{Models and Dataset}
We select Llama-3.1-8B-Instruct \cite{dubey2024llama} and Gemma-2-9b-it \cite{team2024gemma} as our candidate models, chosen for their well-regarded publicly available SAEs. We curated the 1000 prompts from \textit{politically-liberal} subset of \citet{perez2023discovering} using Gemini-2.5 pro \cite{comanici2025gemini} to assign a topic to each of the prompts, resulting in 12 different topics. The curation process is detailed in Appendix \ref{app:data_curation}. The distribution of the topics over these categories is summerized in Table \ref{tab:category-dist}. 


\begin{table}[]
    \centering
    \begin{tabular}{lr}
    \toprule
    Category & Percentage \\
    \midrule
    Social Welfare \& Poverty & 29.9 \\
    Political \& Ideological Stances & 16.9 \\
    Social Equality \& Civil Rights & 15.6 \\
    Healthcare & 7.0 \\
    Climate \& Environment & 5.8 \\
    Tax Policy & 5.5 \\
    Traditional Values \& Gender Roles & 4.7 \\
    LGBTQ+ Rights & 4.4 \\
    Abortion Rights & 3.7 \\
    Immigration \& Refugees & 2.6 \\
    Corporate \& Economic Regulation & 2.4 \\
    Military \& Defense Spending & 1.0 \\
    Gun Control & 0.5 \\
    \bottomrule
    \end{tabular}
    \caption{Distribution of the curated dataset across 12 identified topics.}
    \label{tab:category-dist}
\end{table}

We randomly select 126 prompts for evaluation using stratified sampling, 
ensuring at least 3 prompts from each category. The remaining 874 prompts are used for training the steering vectors.
A Factor Analysis (FA) was performed using the responses to ensure that they primarily scale in one dominant dimension. FA confirms that the data contains a first dominant dimension that accounts for $\sim 40\%$ of the total variance (Appendix \ref{app:factor_analysis}).

\subsection{Steering Model Behaviour}

We investigate the extent to which models can be steered ideologically using their existing knowledge, without finetuning. We employ two primary steering approaches: prompt engineering and activation steering.

\par \textbf{Prompt Engineering}: 
We evaluate model responses under 9 prompting conditions that vary the instructed persona (original, liberal, conservative) and argumentative context (none, liberal arguments, conservative arguments). Details of the prompting conditions and persona instructions are in Appendix \ref{app:prompt_cond} and \ref{app:persona_prompts}. We used Llama-3.3-70B-Instruct to generate supporting and counter arguments for each evaluation statement. We label the responses as $liberal=1$ or $conservative=0$ by matching them with politically liberal answers from Anthropic's dataset. We label the cases where the models refuse or do not provide an answer as a $refusal=null$ response

\begin{figure}[h]
\centering
    \includegraphics[width=\linewidth]{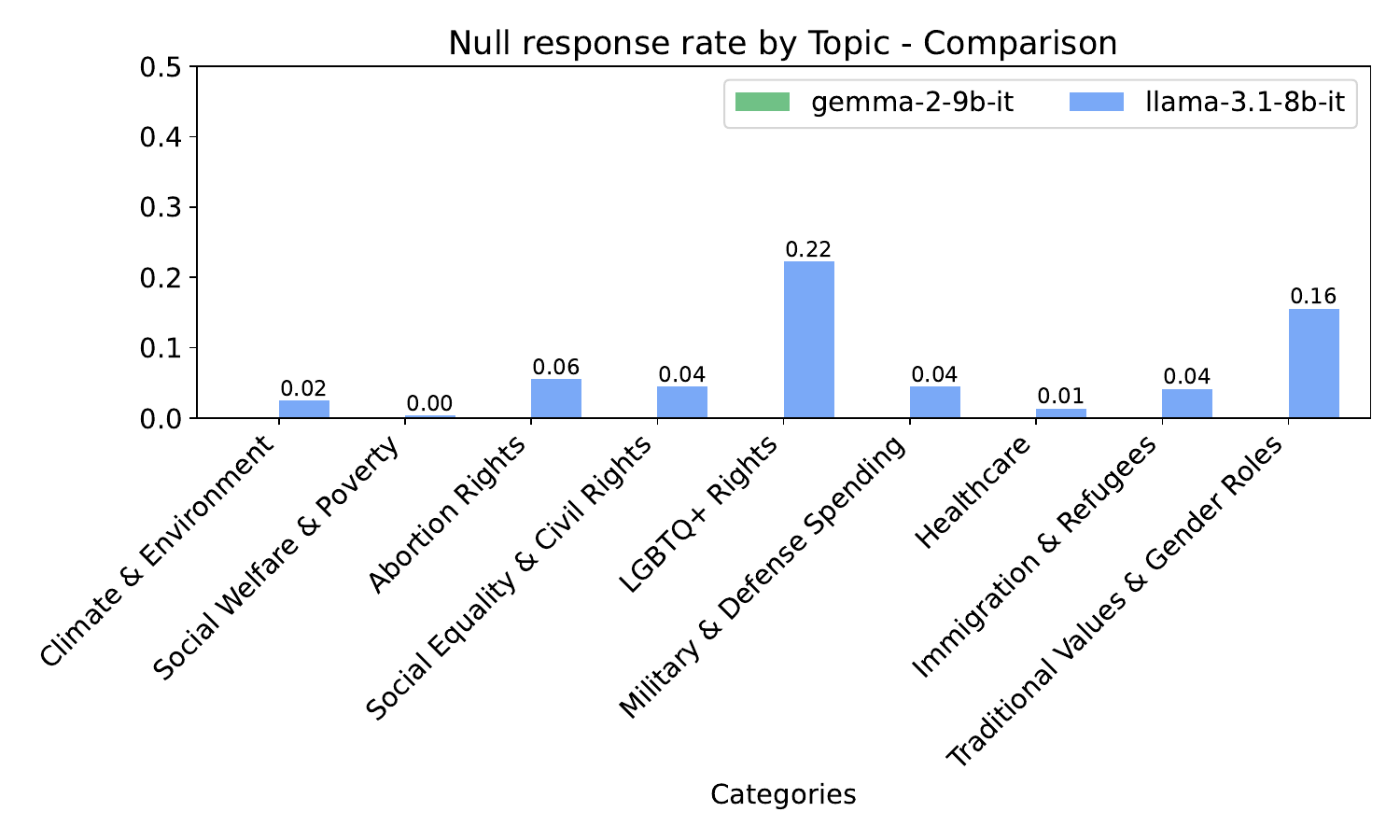}
    \caption{Topic-wise refusal (“null”) rates. Llama refuses more often on socially contentious issues, while Gemma remains responsive.}
    \label{fig:null_rate}
    \vspace{-10pt}
\end{figure}

Our analysis reveals several key findings. First, both models prioritize explicit persona instructions over argumentative pressure—instructing them to answer as a conservative is more effective than providing conservative arguments. Second, Gemma produces more conservative responses than Llama overall and handles both personas more readily. In contrast, Llama frequently refuses to answer when instructed to adopt a conservative stance, particularly on socially contentious topics. Figure \ref{fig:null_rate} shows the models' $null$ response rate over topics or categories. Third, both models exhibit lower response consistency (higher variance) when adopting conservative personas compared to liberal ones, with the least consistency occurring when conservative arguments are presented to the original persona. Detailed topic-level analysis is in Appendix \ref{app:response_cons}

\begin{figure}[!h]
    \centering
    \includegraphics[width=\linewidth]{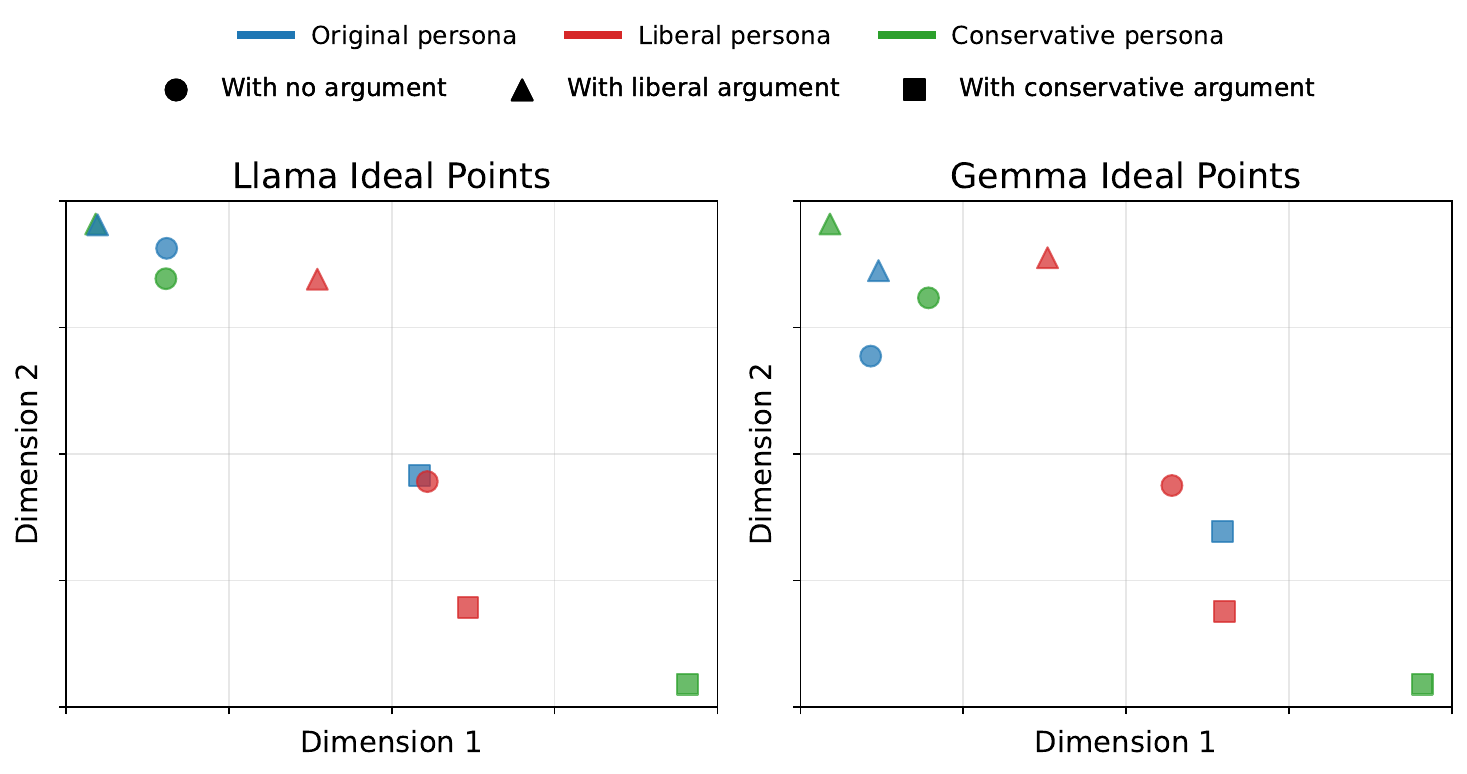}
    \caption{Estimated ideological positions across nine prompting conditions. Variance increases under conservative personas or arguments, revealing unstable conservative representations.}
    \label{fig:model_stance}
    \vspace{-10pt}
\end{figure}

We further analyze the ideological position of the LLMs using the IDEAL method \cite{clinton2004statistical}, which applies Multidimensional Item Response Theory (validation in Appendix \ref{app:val_mirt}). This approach is able to model Llama's selective refusal responses. Figure \ref{fig:model_stance} shows the estimated ideological positions across conditions. The greatest variance in ideological positions occurs under conservative instruction or arguments, suggesting unstable internal representations for conservative-aligned responses. Given the higher variance and richer feature diversity observed in the conservative role, we focus subsequent experiments on steering toward conservative responses.

\par \textbf{Activation Steering}: 

We compare two activation steering methods: (1) Contrastive Activation Addition (CAA) \cite{rimsky-etal-2024-steering}, and (2) Steering Target Atoms (STA) \cite{wang2025promptengineeringrobustbehavior}, which selects target features from SAE representations based on activation amplitude and frequency. We use SAEs \textit{14-llamascope-res-131k} and \textit{20-gemmascope-res-131k} from LlamaScope \cite{he2024llamascopeextractingmillions} and GemmaScope \cite{lieberum-etal-2024-gemma}, respectively. Although LlamaScope provides SAEs only for the base model, prior work suggests they remain effective for instruction-tuned model variants \cite{rimsky-etal-2024-steering}. Layer selection was determined through layer sweeps (Appendix \ref{app:layer_sweep}), identifying layers 14 and 20 as optimal for Llama and Gemma, respectively.

\begin{figure}[h]
    \centering
    \includegraphics[width=\linewidth]{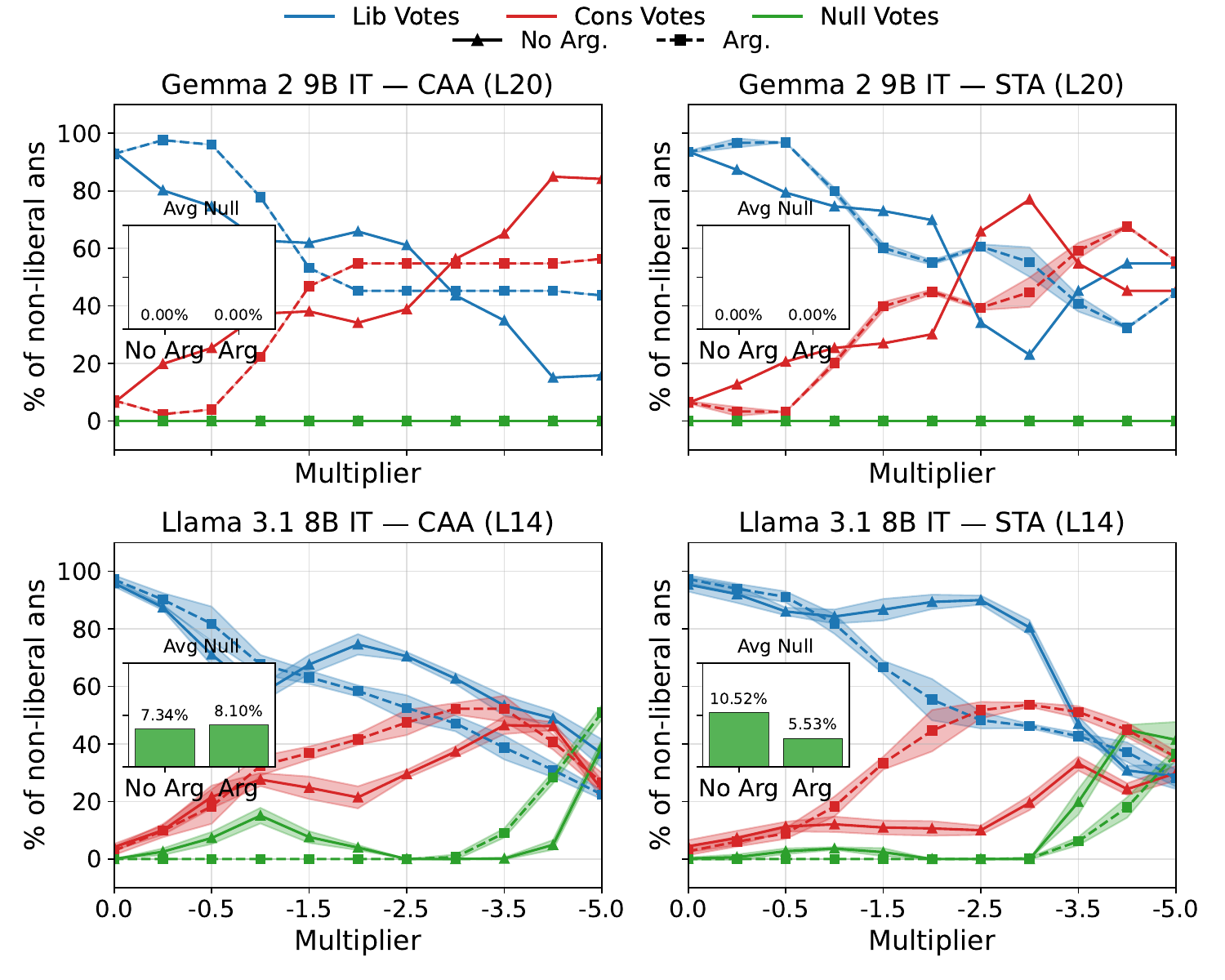}
    \caption{Effects of Contrastive Activation Addition (CAA) and Steering Target Atoms (STA) on response polarity. Gemma shifts smoothly toward conservative answers; Llama mostly increases refusals. Lines show the mean of 5 trials; shaded regions represent the 95\% confidence interval.}
    \label{fig:steering_effect}
\end{figure}

Figure \ref{fig:steering_effect} shows the steering effects. Both CAA and STA methods successfully shift models toward conservative responses, but with striking differences. Gemma steadily increases conservative responses while resulting in minimal refusals. Conversely, Llama predominantly increases refusals rather than conservative answers, suggesting it lacks the internal features necessary to generate conservative responses on many topics. This pattern holds under argumentative pressure.

\begin{figure}[h]
    \centering
    \includegraphics[width=\linewidth]{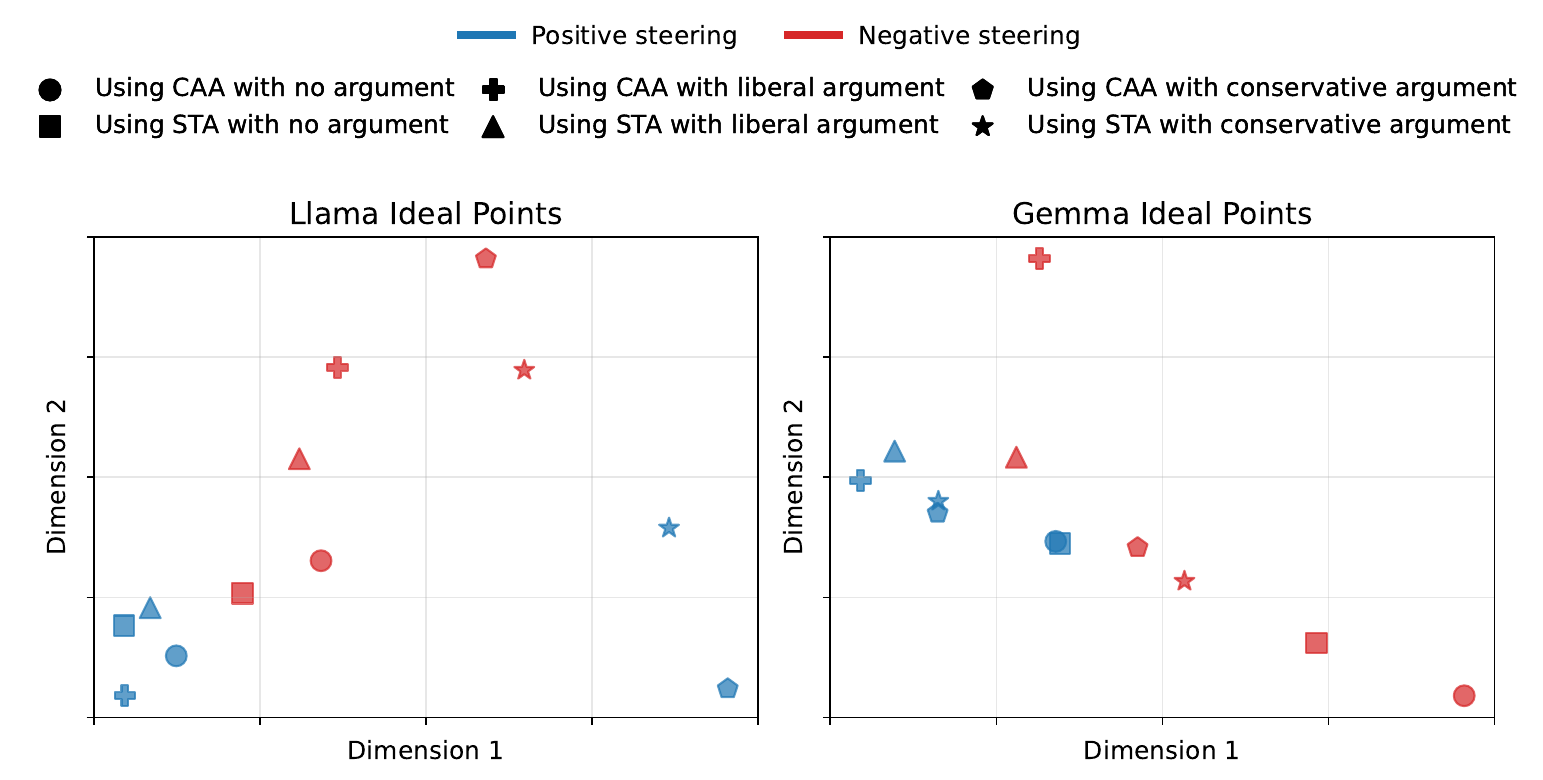}
    \caption{Ideal-point estimates under activation steering with multiplier +1 and -1 corresponding to liberal and conservative steering, respectively. Gemma forms distinct clusters for liberal vs. conservative directions; Llama’s overlap indicates weaker separation.}
    \label{fig:model_stance_steered}
    \vspace{-10pt}
\end{figure}

The ideal points of the models after steering them under argumentative conditions also show stark differences, as shown in Figure \ref{fig:model_stance_steered}. Gemma consistently results in more variance in their responses for the same conditions, compared to Llama, as can be seen by the distance between blue and red point pairs. This results in clearer non-overlapping clusters due to +ve and -ve steering. Additional topic-level consistency analysis is provided in Appendix \ref{app:response_cons}.

\subsection{Political Feature Analysis}
Motivated by these findings, we turned our attention towards analyzing the Sparse Autoencoder (SAE) features, hoping to find a difference in their political feature that can explain their behavioral differences while acting as a liberal or a conservative.

SAEs achieves this by projecting the vector to a high-dimensional vector space with the goal of disentangling the complex features that are otherwise superpositioned in the outputs of a single neuron \cite{elhage2022toymodelssuperposition, bricken2023monosemanticity}. 
Our goal was to see if the SAE features from these layers can be used to quantify their political knowledge and subsequently explain why their capabilities vary when playing different roles.
Although there are typically thousands of features in a single SAE, not all of them activate or are equally important for a particular behavior. Recent SAE-based steering approaches select features based on their activation values (amplitude), activation frequencies \cite{wang2025promptengineeringrobustbehavior}, and their steering strength \cite{arad2025saesgoodsteering}. First, we conduct a \textit{quantitative} comparison to measure the raw number and activation strength of activated features in each model. We then perform a \textit{qualitative} evaluation to assess the thematic coherence and predictive power of these features, to confirm how they represent genuine political concepts.

\par \textbf{Quantitative Analysis:} We used the approach of \citet{wang2025promptengineeringrobustbehavior} to collect and analyze the normalized mean amplitude (activation difference) and the normalized mean frequency difference of the activated features for positive and negative data from the 874 training prompts using  $\Delta a = \frac{1}{N}\sum^N_i (\bar{a}_i^{pos}-\bar{a}_i^{neg})$ and $\Delta f = f^{pos} - f^{neg}$. Where, $f^{pos,j} = \frac{1}{N}\sum^N_i \mathbb{I} (|\bar{a}_i^{pos,j}|>0)$ and $f^{neg,j} = \frac{1}{N}\sum^N_i \mathbb{I} (|\bar{a}_i^{neg,j}|>0)$.

For LLama, we found 167 features with $\Delta a>0$ and 157 features with $\Delta f>0$. In contrast, in Gemma we found 368 and 334 features with $\Delta f>0$ and $\Delta a>0$, respectively. 

\begin{table}[h]
\centering

\begin{tabular}{lll}
\toprule
\textbf{} & \multicolumn{1}{c}{Gemma-2-9b-it} & \multicolumn{1}{c}{Llama-3.1-8b-it} \\
\midrule
Mean               & 0.002407   & 0.018038   \\
Std. Dev.          & 0.041598   & 0.059238   \\
Min.               & 0.000000   & 0.000000   \\
Q1                 & 0.000000   & 0.000001   \\
Median             & 0.000000   & 0.000003   \\
Q3                 & 0.000002   & 0.009067   \\
Max.               & 0.882730   & 0.247728   \\
\bottomrule
\end{tabular}
\caption{Summary statistics of SAE feature output scores. Gemma's features consist of a few extremely high-quality features that pull the mean above the median despite most values being clustered near zero.}
\label{tab:output_score_stats}
\end{table}

We also collected all the features that activate for the 126 statements in our evaluation set by projecting the activation vector of the target layer on the SAE space, marking the token of the statement that most activates this feature, and neglecting features activated for the bos token. In total, for the gemma-2-9b-it model, we found 18458 features, whereas for the llama-3.1-8b-it, we found only 4412 activated features out of a total possible 131K features in both SAEs. On average, Gemma contained \textbf{7.3$\times$ more features} that activate for a statement. 

Next, we selected features with $\Delta f>0$ or $\Delta a>0$ from both models and evaluated the quality of these features based on their ability to steer the model's behaviour. We got 71 and 16 features from Gemma and Llama SAEs, respectively. This is done because if the features can steer the model heavily, even a small number of features can be enough to control the model. To do this, we calculated the output scores of the features, using the approach proposed by \citet{arad2025saesgoodsteering}, by performing interventions during a forward pass and evaluating the change in the rank and probability assigned to tokens in the target layer. The output score is calculated using Equation \ref{eq:output_score}, which is the difference between the original and the interveined rank weighted probabilities of the target layer when the model is provided with a neutral sentence (we used "In my opinion,"). $r(l^*, \mathcal{M})$ and $p(l^*, \mathcal{M})$ denote the rank and probability of the highest-ranked token $l^*$, and $|V|$ is the vocab size of the model $\mathcal{M}$. 
\begin{equation}
    \begin{split}
        P(\mathcal{M}) = (1-\frac{r(l^*, \mathcal{M})}{|V|})p(l^*, \mathcal{M}), \\
        S_{out}=P(\mathcal{M}_{h\leftarrow{\phi(h)}}-P(\mathcal{M}).
    \end{split}
    \label{eq:output_score}
\end{equation}

The stats of the output scores are shown in Table \ref{tab:output_score_stats}. The scores for Gemma appear to be heavily right-skewed, with most values clustered near zero. The median and even the 75th percentile (Q3) are $zero$. However, its maximum score reaches 0.88. This suggests Gemma contains a few extremely high quality features, pulling the mean (0.002407) above the median. In contrast, the maximum feature score for Llama was significantly lower.

\par \textbf{Qualitative analysis:}
To assess the quality of the features, we use GPT-4o-mini \cite{hurst2024gpt} as an automated evaluator to analyze the coherence and predictive validity of SAE features from each model. We make sure to ask the LLM to provide explanations justifying each evaluation, a technique shown to improve alignment with human judgments \cite{chiang-lee-2023-closer}. First, we collected the feature descriptions of the features of the SAE with $\Delta a$ or $\Delta f$ from Neuronpedia\footnote{https://www.neuronpedia.org/}. Next, we use the prompts in Appendix \ref{app:quality_prompts}, to perform two distinct, complementary evaluations:

\begin{enumerate}
    \item Predictive Validity: To test the practical utility of the features, we tasked the evaluator with classifying the original statement into one of the predefined 12 political categories based solely on the provided feature description list. For each statement, the evaluator returned the chosen category and a confidence score on a 1-5 scale. 
    
    \item Thematic Coherence: To assess the interpretability and theoretical soundness of the features, we conducted a second evaluation. Here, the evaluator was given \textbf{only} the list of activated features' descriptions and asked to rate their thematic coherence on a 1-5 Likert scale, where 1 represented a disjointed, random set and 5 represented a highly unified concept.    
\end{enumerate}

\begin{figure}[t]
    \centering
    \includegraphics[width=\linewidth]{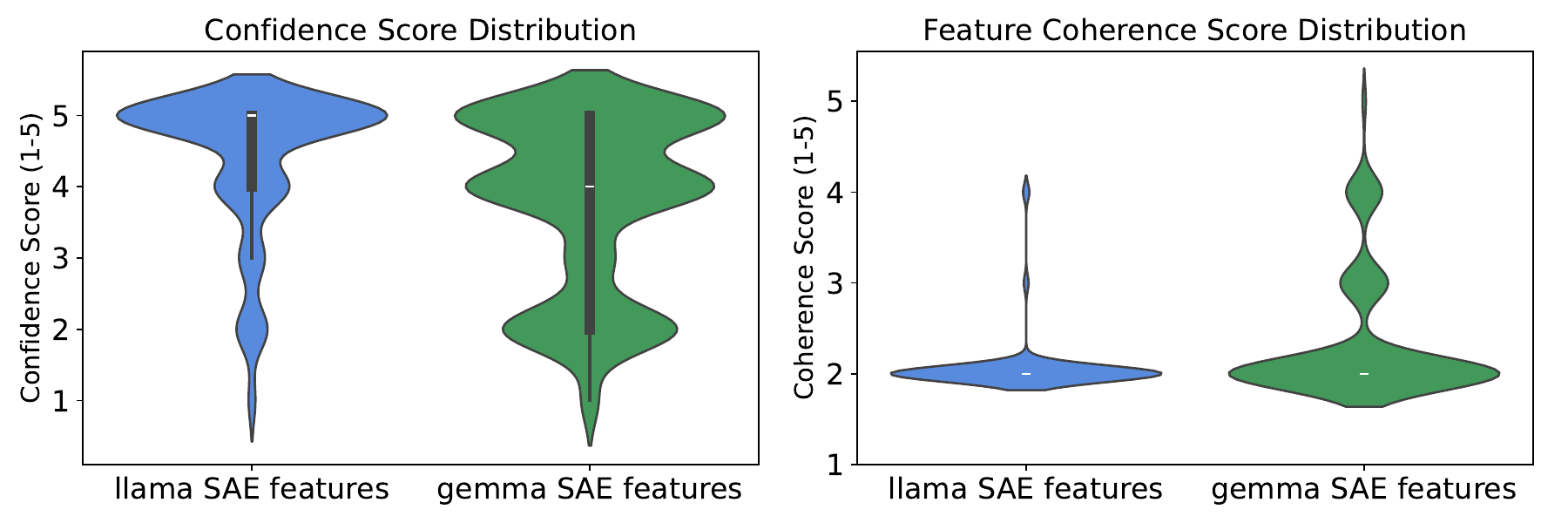}
    \caption{Distributions of feature evaluation scores. Gemma’s SAE features achieve higher thematic coherence (right) and stronger predictive validity than Llama’s (left).}
    \label{fig:feat_conf_coh}
\end{figure}

Our evaluation reveals significant differences in the SAE features from  llama-3.1-8b-it and gemma-29b-it in terms of quality and utility for modeling political contexts. Figure \ref{fig:feat_conf_coh} provides an initial, high-level overview of feature quality based on the two key scores: classification confidence and thematic coherence. The "Confidence Score Distribution" (left) shows that the evaluator was able to make high-confidence classifications using features from both models. However, a confidence score of 5 mostly turned out to be where the evaluator could not assign any category, i.e., a confident failure. These “confident failure” outcomes made up 69\% for Llama and 39\% for Gemma. Beyond this, Gemma’s scores cluster more tightly around 4 and 2 (cases where a category was successfully mapped), while Llama’s distribution has a longer tail toward lower scores, indicating less predictive clarity.
The "Feature Coherence Score Distribution" (right) reveals a more dramatic distinction. Llama's features are consistently rated as having low coherence, with scores tightly clustered around a median of 2. This indicates that its feature sets are frequently perceived as a disjointed mix of unrelated concepts. In contrast, Gemma's coherence scores are not only higher on average but also show a wider distribution, with an additional concentration of scores between 3.0 and 5.0. This indicates that Gemma is capable of generating highly coherent and interpretable feature sets that represent a unified political or social theme.

\begin{figure}[h]
    \centering
    \includegraphics[width=\linewidth]{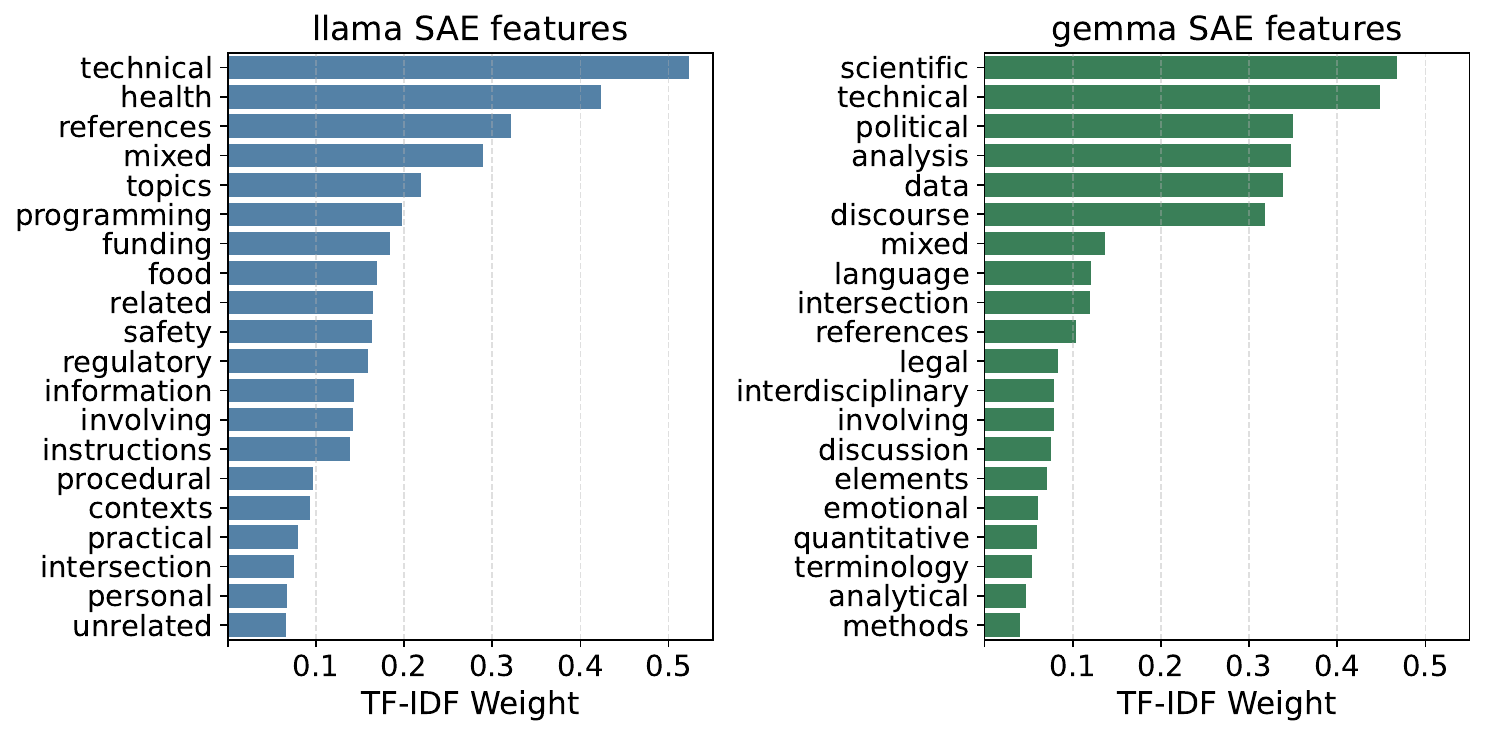}
    \caption{TF-IDF–weighted term distributions from SAE feature descriptions. Gemma emphasizes politically and analytically rich terms; Llama focuses on procedural or structural words.}
    \label{fig:theme_tf_idf}
    \vspace{-5pt}
\end{figure}

The term frequency plots in Figure \ref{fig:theme_tf_idf} provide a quantitative view of the thematic gap between the feature sets. The Llama features are dominated by general, structure-oriented terms such as technical, health, references, and procedural, indicating a focus on syntactic and procedural aspects rather than on semantic or ideological content. Although domain terms like funding and safety appear, they remain embedded in practical rather than interpretive contexts—consistent with lower coherence scores and a descriptive rather than analytical framing. In contrast, Gemma’s features emphasize scientific, political, and analytical discourse, with frequent terms such as political, analysis, data, and discourse highlighting a richer engagement with evaluative and interpretive language. This concentration reflects stronger alignment with socially and politically grounded semantics.

\begin{figure*}[t]
    \centering
    \includegraphics[width=\linewidth]{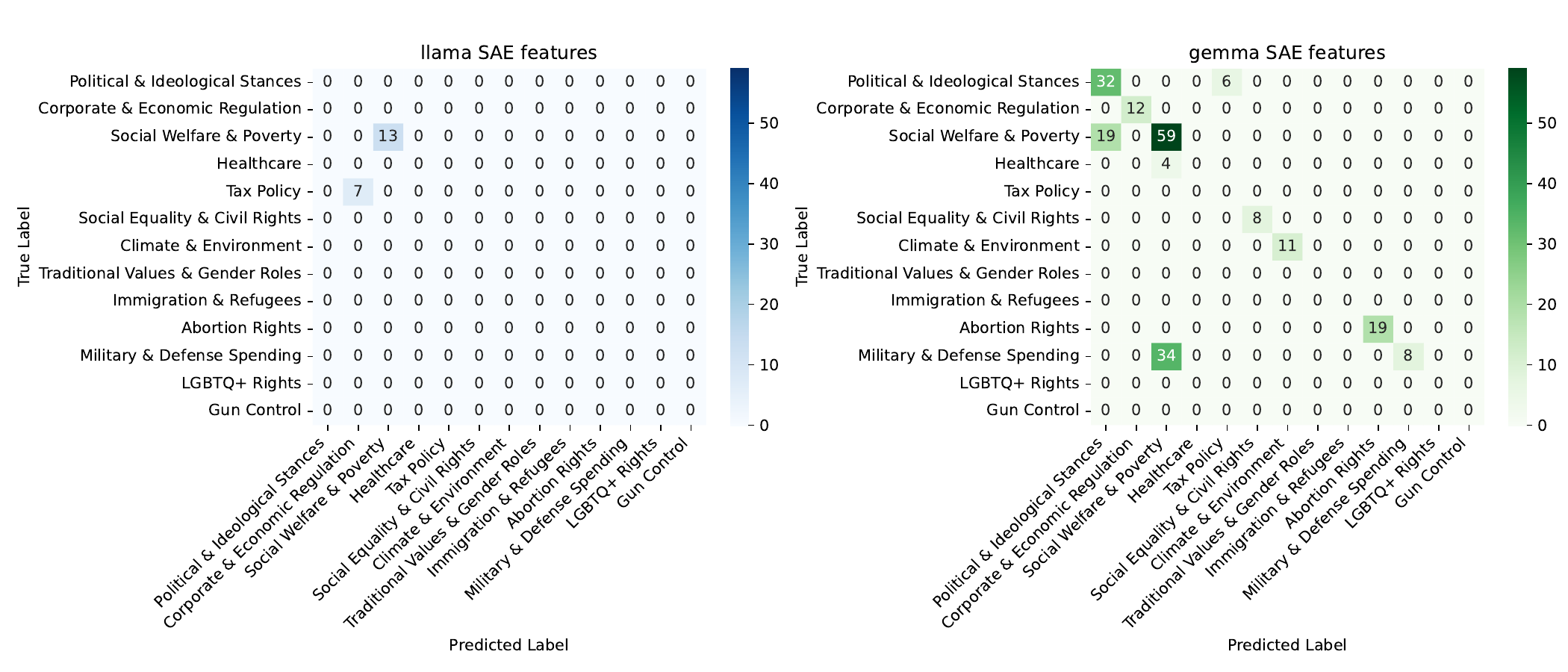}
    \caption{Confusion matrices of topic classification using SAE feature descriptions. Gemma’s features yield clear diagonals and topic discrimination; Llama’s fail to separate categories.}
    \label{fig:feat_confusion_mat}
    \vspace{-10pt}
\end{figure*}

Finally, the confusion matrices in Figure \ref{fig:feat_confusion_mat}  using the categories of the statements as ground truth and the categories identified in the 1st experiment provide definitive, quantitative evidence of how these differences in feature quality translate to performance on a practical downstream task. The matrix for Llama-3.1-8b-it shows a near-total failure to classify statements correctly. 
Aside from a small number of successful classifications in "Social Welfare \& Poverty" and "Tax Policy", the model's features provide insufficient signal to distinguish between the nuanced political categories. The matrix for Gemma-2-9b-it, however, demonstrates a strong and effective classification capability. The pronounced diagonal line indicates a high number of true positives across a wide range of complex topics, including "Political \& Ideological Stances" (32), "Social Welfare \& Poverty" (59), "Climate \& Environment" (11), "Abortion Rights" (19), and "Military \& Defense Spending" (34). While some misclassification exists, for instance, misclassifying "Social Welfare" statements as "Political Stances", these are semantically plausible. The overall result confirms that the high coherence and domain-specific nature of Gemma's features provide a robust basis for accurate political text classification, a task for which Llama's more generic and less coherent features are demonstrably inadequate.


\subsection{Ablation of Political Features}
\begin{figure}[h]
    \includegraphics[width=\linewidth]{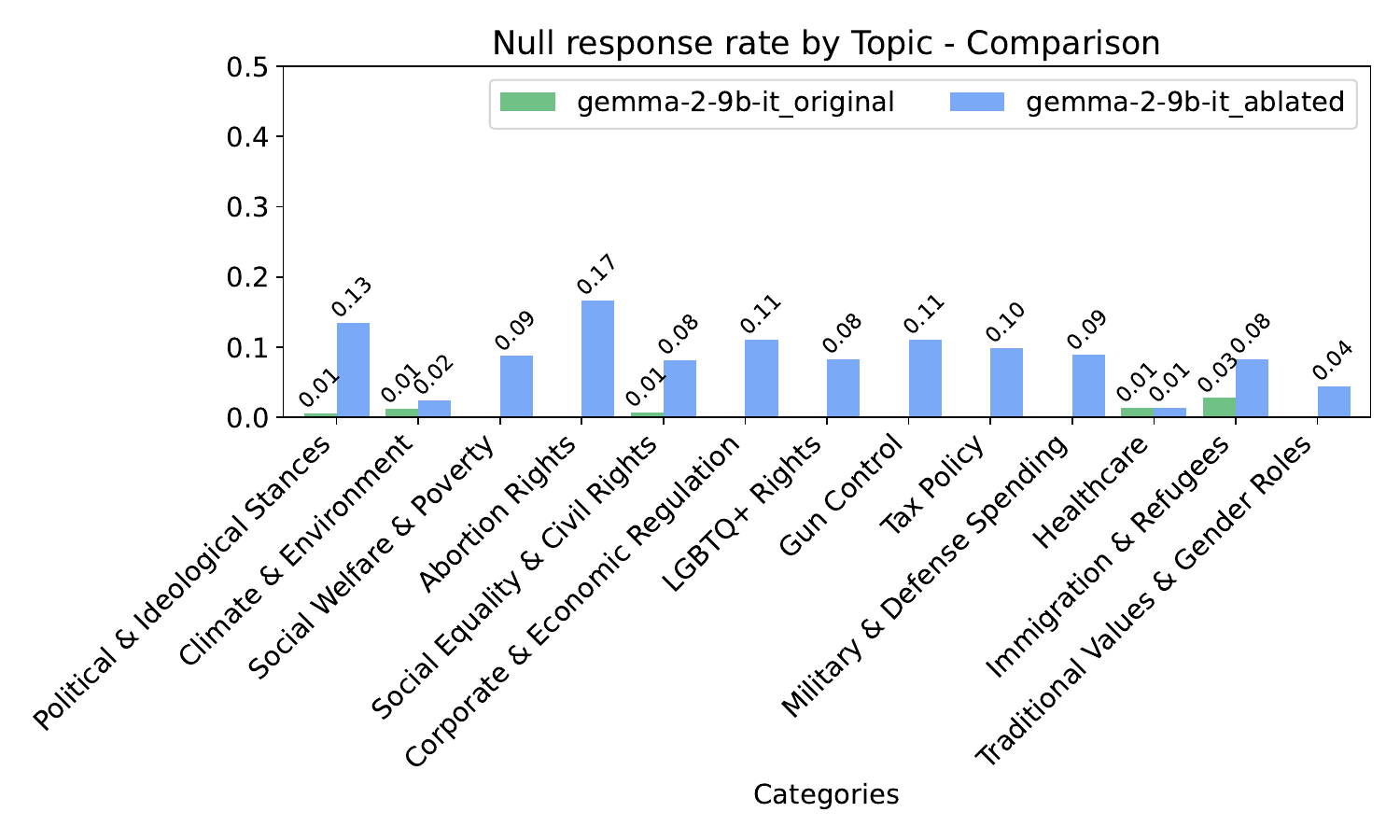}
    \caption{Refusal rates before and after ablation of Gemma’s top political features. Removing these features sharply increases refusals, linking capability loss to ideological shallowness.}
    \label{fig:null_rate_ablated}
    \vspace{-10pt}
\end{figure}

To directly test the causal link between political features and the refusal behavior, we performed a zero-ablation of the 71 most salient political features in Gemma-2-9b-it and evaluated their null response rate under the argumentative setting like before. The ablation caused a sharp increase in the model's refusal rate across most topics, as shown in Figure \ref{fig:null_rate_ablated}. 
The ablated model, which originally had a negligible refusal rate, now frequently defaulted to generic safety-aligned responses or failed to follow instructions (though this may be more due to ablation removing some capability, but this was less frequent). This result provides strong causal evidence that refusal can arise from a capability deficit. Removing the specific features that encode political concepts appears to render the model unable to reason about the topic from the instructed perspective, forcing a fallback on its general safety training and mirroring the brittle behavior observed in the feature-poor Llama model.

\section{Discussion}
Our findings reveal that ideological depth in LLMs is a measurable structural property reflected in their internal feature representations. Comparing Llama-3.1-8B-IT and Gemma-2-9B-IT, we demonstrate that models exhibit vastly different political feature spaces: Gemma possesses $\sim7.3\times$ more political features than Llama, directly explaining their behavioral differences. Gemma's rich feature space enables flexible reasoning across ideological perspectives and high steerability, while Llama's sparse features result in brittleness and frequent refusals.

Critically, we identify that Llama's refusal behavior stems from a \textit{capability deficit} rather than sophisticated safety guardrails. This has important implications for AI alignment. An ideologically shallow model may appear safer through refusal, but this brittleness creates unpredictability -- its refusal reflects a failure of understanding, not principled reasoning. Conversely, a feature-rich model like Gemma can more readily adopt undesirable personas, resulting in a more transparent and interpretable behavior, offering better opportunities for monitoring and intervention.

We validate the semantic meaningfulness of these features through automated topic classification using only SAE feature descriptions. The evaluator's success with Gemma's features (achieving strong classification performance) confirms they form a coherent political knowledge map, while its failure with Llama's generic features corroborates our finding of shallow ideological representations.

\section{Conclusion and Future Work}

We demonstrate that ideological depth in LLMs is measurable through SAE feature analysis and directly linked to model steerability. Our comparison of Llama-3.1-8B-IT and Gemma-2-9B-IT reveals that Gemma's larger political feature space enables flexible ideological reasoning, while Llama's sparse representations result in brittleness and elevated refusal rates. Moreover, we show that targeted ablation of salient political features induces refusal behavior in Gemma, establishing a \textit{causal link} between feature richness and response capability.

Our future research directions include:
\begin{enumerate}[noitemsep, topsep=3pt]
    \item \textbf{Mechanistic origins of response instability}: Models exhibit significantly higher response variance under conservative personas. Future work should trace the feature circuits responsible for this instability \cite{ameisen2025circuit, lindsey2025biology}.
    \item \textbf{Real-time monitoring}: Developing activation probes for political features to enable interpretability and control during inference \cite{chen2025personavectorsmonitoringcontrolling}.
    \item \textbf{Training interventions}: Investigating whether ideological depth can be deliberately shaped during fine-tuning or pretraining.
\end{enumerate}

Understanding these latent structures is essential for building LLMs that are simultaneously capable, predictable, and alignable.

\section{Limitations}

Our analysis relies on Sparse Autoencoders (SAEs), which have known limitations \cite{smith2025negative}. SAEs heavily depend on training data \cite{SAEsAreHighlyDatasetDependent}, which could confound our comparisons. However, both SAEs used identical sizes (131K features) and were trained on comparable non-chat corpora (pile-uncopyrighted for GemmaScope, SlimPajama-627B for LlamaScope), providing a largely controlled comparison. The difference in feature magnitude and stark contrast in downstream classification performance suggest genuine representational disparities beyond methodological artifacts. Future work will validate these findings using complementary methods such as probing \cite{alain2016understanding}, which has shown superior performance on other interpretability tasks.

Additionally, our study focuses on two models of similar size. Whether our findings generalize to larger models or models from different training paradigms remains an open question. It is possible that at a sufficient scale, all models develop rich political feature representations, or conversely, that scaling exacerbates rather than ameliorates representational disparities.

Finally, while we focus on the political domain due to its societal importance and well-defined ground truth labels, ideological reasoning is multifaceted and culturally situated. Our analysis uses US-centric political categories, which may not capture the full complexity of global political discourse. Future work examining non-Western political frameworks would provide a valuable perspective on the generalizability of our findings.

\bibliography{references.bib}

\newpage
\appendix

\section{Data Curation Process}
\label{app:data_curation}
We used the 1000 prompts from \textit{politically-liberal} subset of \citet{perez2023discovering}, consisting of dichotomous questions, intended to identify politically liberal behavior from agentic models. We used a two-step approach to assign a topic for each of the statements: In step 1, we passed all of the statements as a list and prompted Gemini-2.5 pro to group them in categories. However, this results in the LLM regenerating all the statements, which sometimes modifies some of them, a scenario known as \textit{context rot} \cite{hong2025context}. Therefore, in step 2, we collect all the categories identified by Gemini and select one statement from the original list, and again prompt the LLM to pick the topic that best matches the statement. To account for \textit{selection bias} of the LLMs -- where they are more likely to pick an option based on their position rather than meaning \cite{zheng2024large, rottger2024political} -- we prompt the LLM 10 times for each statement, each time shuffling the list of categories, and choose the most frequent category for each statement. This ensures that the statements are neatly grouped under distinct topics and no statement is missed or modified during categorization.

After this, all 1000 statements were categorized into 12 different topics. 

\section{Prompting Conditions}
\label{app:prompt_cond}
\begin{table}[h]
    \centering
    \begin{tabular}{l}
    \toprule
        Original persona with no arguments \\
        Original persona with liberal arguments \\
        Original persona with conservative arguments \\
        Liberal persona with no arguments \\
        Liberal persona with liberal arguments \\
        Liberal persona with conservative arguments \\
        Conservative persona with no arguments \\
        Conservative persona with liberal arguments \\
        Conservative persona with conservative arguments \\
    \bottomrule
    \end{tabular}
    \caption{Prompting conditions used.}
    \label{tab:promt_eng}
    \vspace{-10pt}
\end{table}

Along with the LLMs' response to the prompts as their original personas, we also provide detailed instructions to the model to answer the prompts as an individual with either a liberal or a conservative worldview (Appendix \ref{app:persona_prompts}). Additionally, to analyze the ideological depths of the LLMs while answering as different personas, we also evaluate stance consistency in the presence of supporting and counter arguments in the prompt \cite{kabir2025wordsreflectbeliefsevaluating}. We used Llama-3.3-70B-Instruct to generate at least 5 different supporting and counter arguments for each of the 126 statements in our evaluation set. Combining all these, our approach analyzes 9 conditions using prompt engineering, which are shown in Table \ref{tab:promt_eng}.

\section{Response Consistency}
\label{app:response_cons}

We also analyze the response consistency over the 9 prompting conditions by calculating the variance of their response over each condition. Here $ consistency = 1-4 \times var(responses) $ where 4 is a scaling factor, to keep the values in the range $[0,1]$, where maximum variance is 0.25 ($consistency=0$). Both models' consistency was in agreement. 
\begin{figure}[h]
    \centering
    \includegraphics[width=\linewidth]{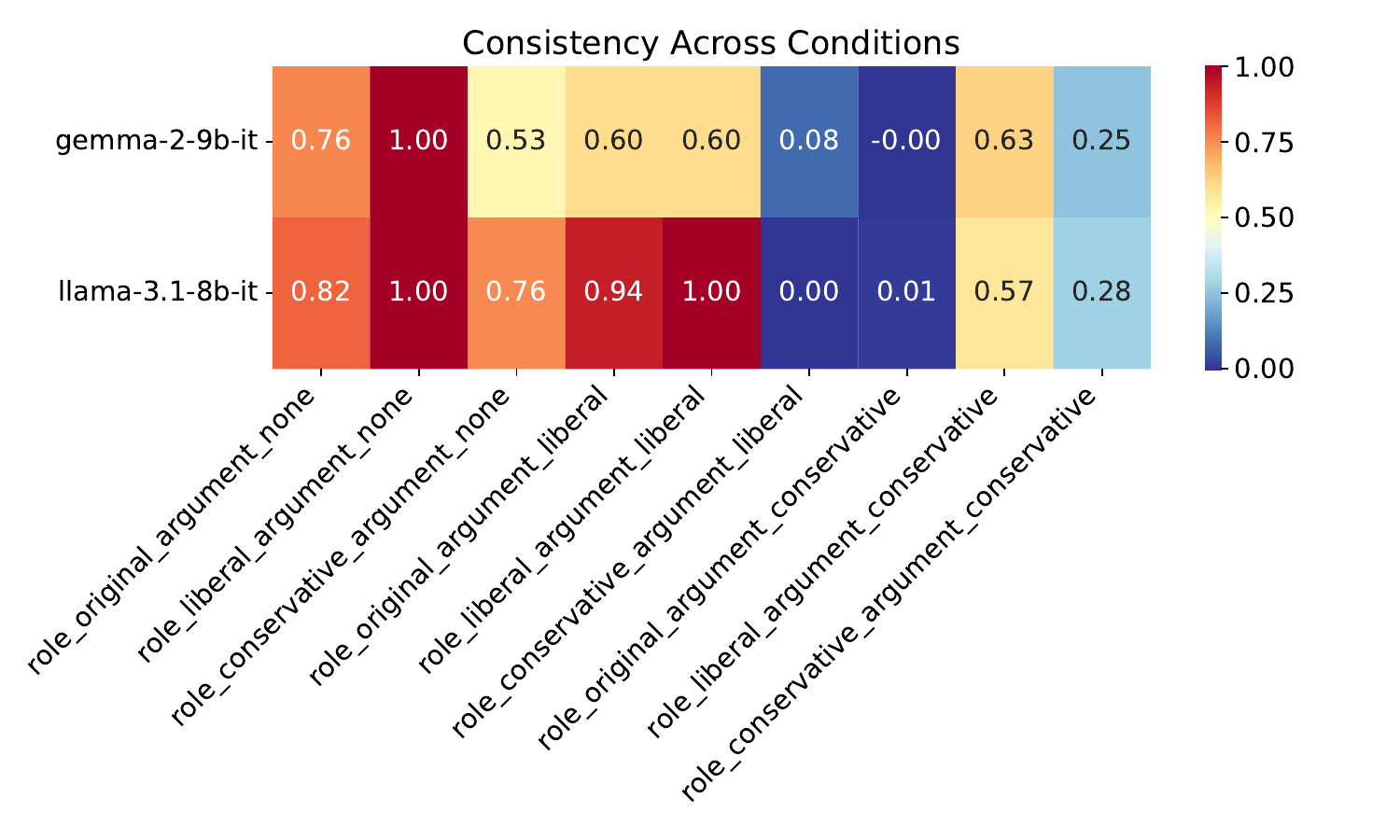}
    \caption{Heatmap of response consistency across nine prompting conditions. Models show the least consistency (most variance) when adopting conservative roles or arguments.}
    \label{fig:response_var}
    \vspace{-10pt}
\end{figure}
As shown in Figure \ref{fig:response_var}, both models were more consistent when instructed to answer as a liberal, even compared to answering as their original personas, and were less consistent when answering as a conservative. Models were least consistent when dealing with conservative arguments while playing their original persona, and liberal arguments when playing the conservative persona. This means instructions have a significant impact on how well the models deal with arguments. 
As the models show more inconsistency while acting as a conservative, the rest of the experiments only concern their differences when playing the conservative role. We also find more diverse features while playing the conservative role in their SAEs.

\begin{figure}[h!]
    \centering
    \begin{subfigure}[t]{\linewidth}
        \centering
        \includegraphics[width=\textwidth]{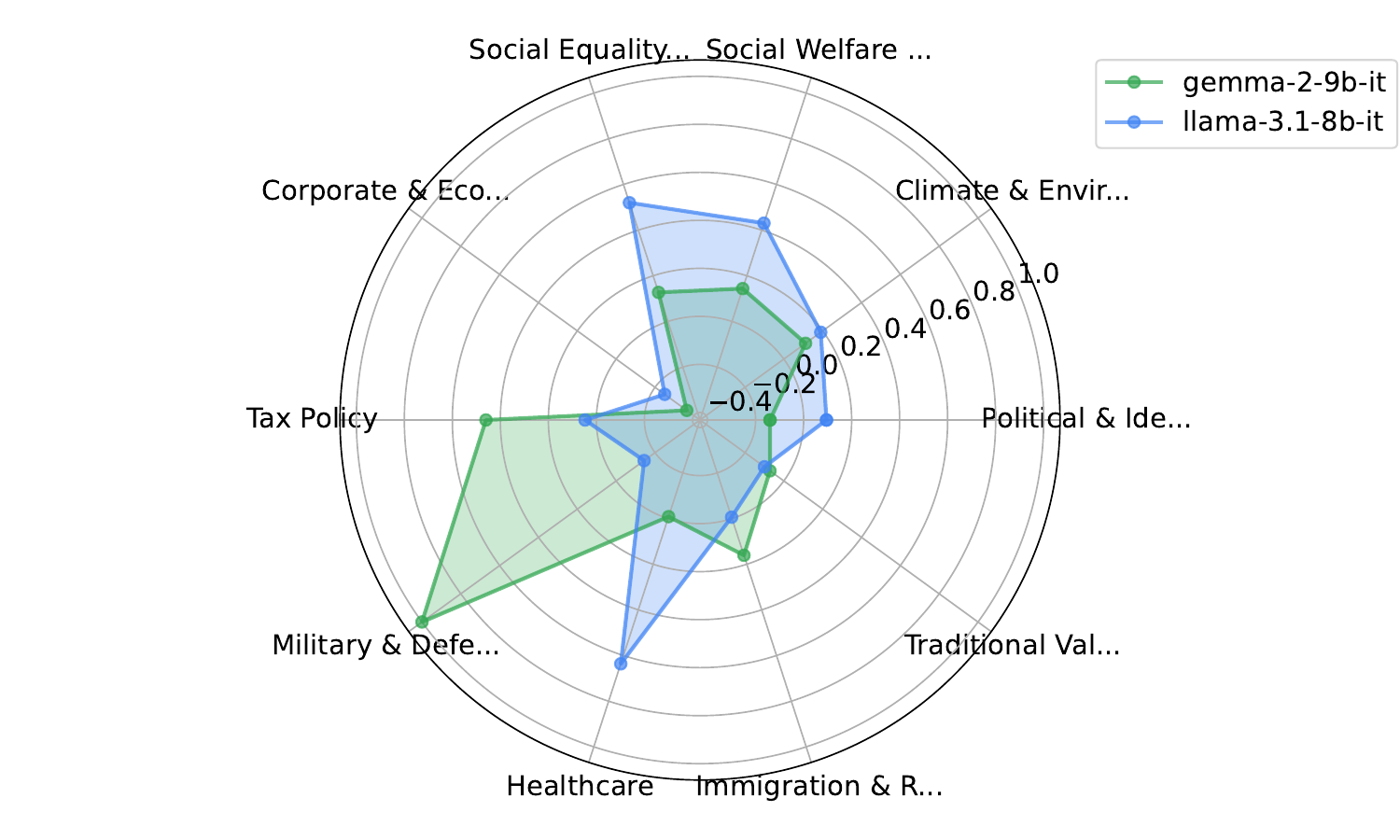}
        \caption{Response consistency Fleiss' $\kappa$ over topics.}
        \label{fig:response_coh}
    \end{subfigure}
    \hfill
    \begin{subfigure}[t]{\linewidth}
        \centering
        \includegraphics[width=\textwidth]{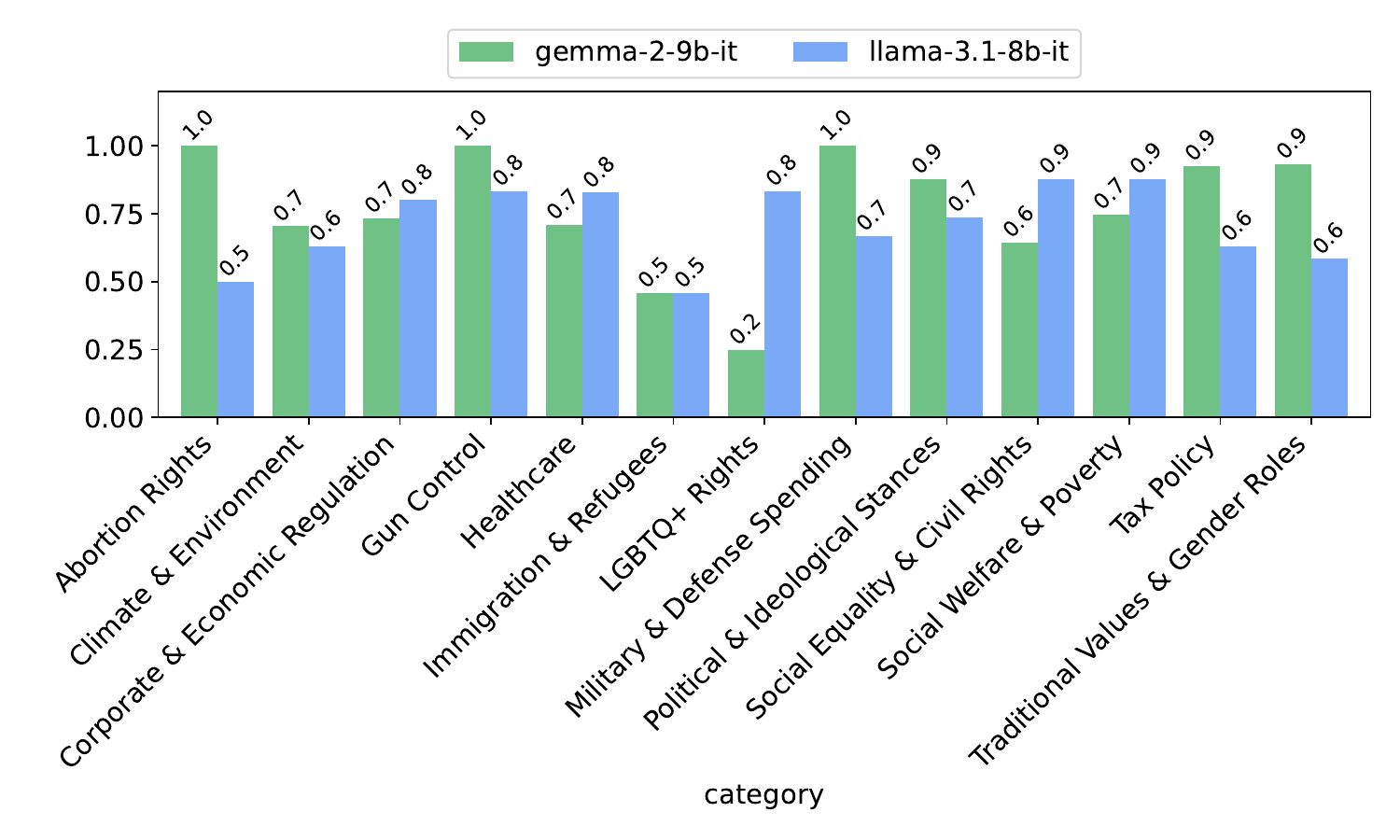}
        \caption{Conservative response tendency.}
        \label{fig:response_tend}
    \end{subfigure}
    \caption{Topic-level conservative consistency and tendency under argumentative pressure when instructed to adopt a conservative persona. Gemma achieves stronger conservative alignment but with greater response variance than Llama.}
    \vspace{-10pt}
\end{figure}

Next, to gauge the political leaning of each model, we evaluated its response coherence measured by Fleiss' $\kappa$ (Figure \ref{fig:response_coh}) and its conservative response rate (Figure \ref{fig:response_tend}) for each topic when instructed to respond as a conservative under argumentative pressure.

\begin{figure*}[h]
\centering
    \begin{subfigure}[b]{0.48\textwidth}
    \includegraphics[height=0.48\textwidth]{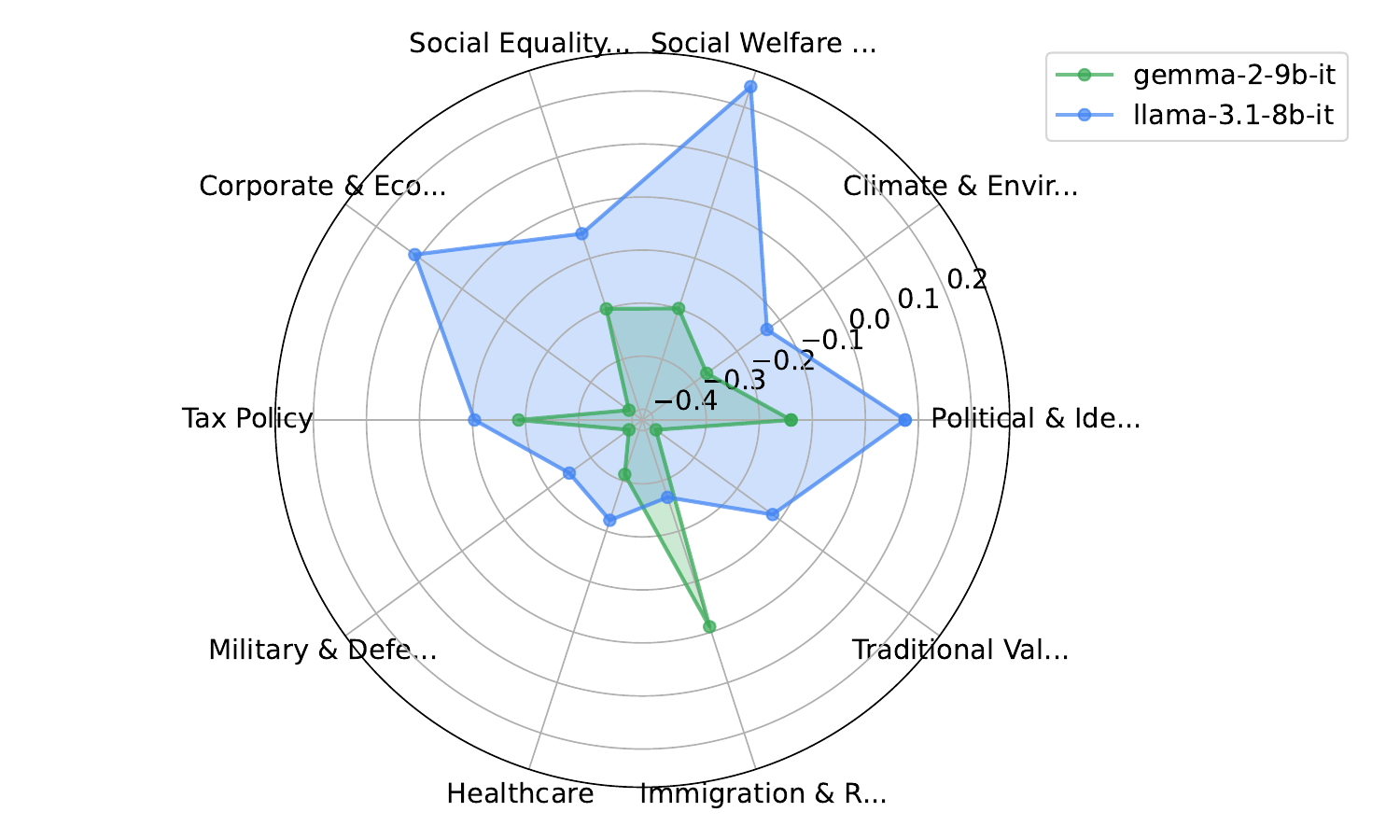}
    \caption{CAA Steering}
    \end{subfigure}
    \begin{subfigure}[b]{0.48\textwidth}
    \includegraphics[height=0.48\textwidth]{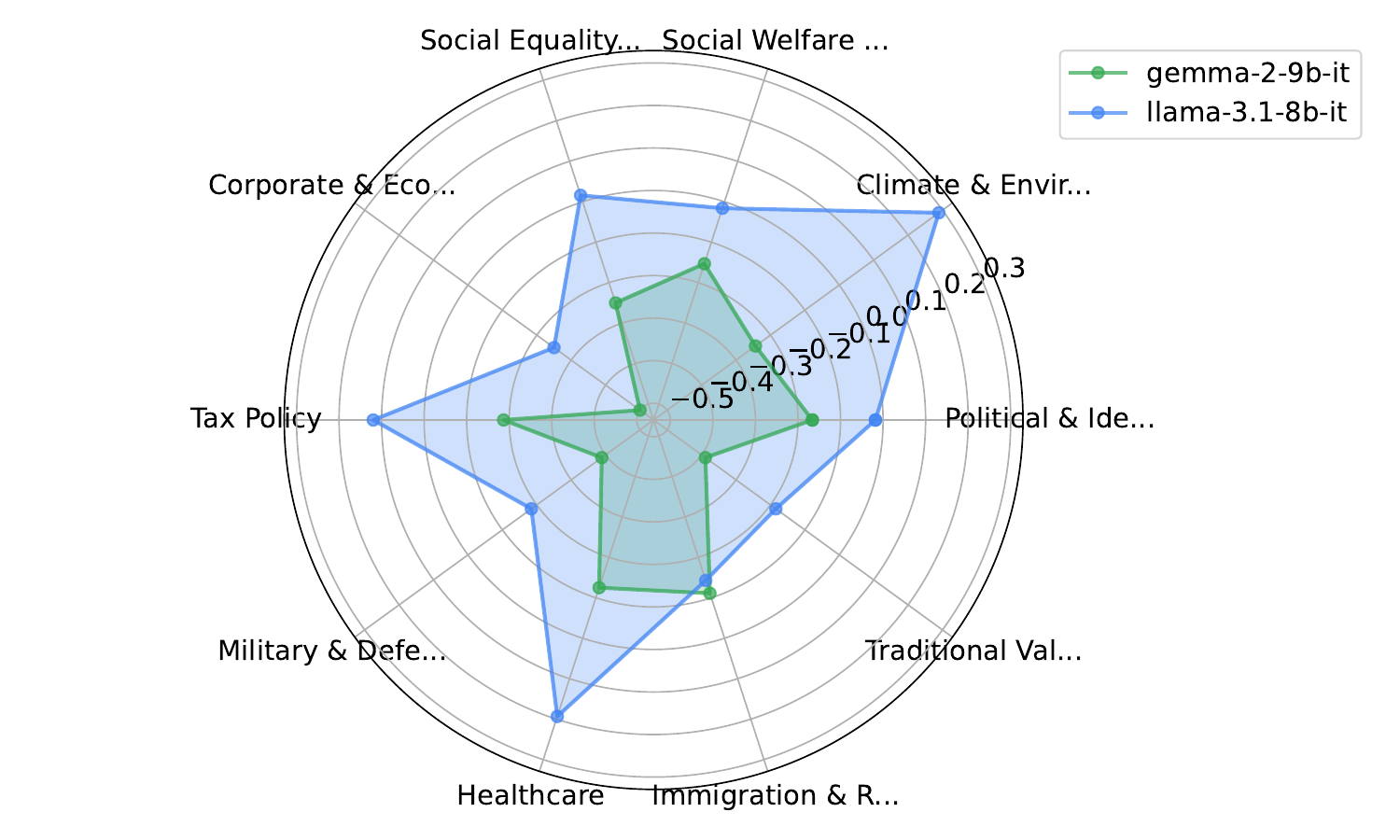}
    \caption{STA Steering}
    \end{subfigure}
    \caption{Topic-level response consistency (Fleiss' $\kappa$) under activation steering toward the conservative direction ($\text{multiplier}=-1$) with argumentative prompts. Gemma achieves larger conservative shifts but lower consistency.}
    \label{fig:model_coh_steer}
    \vspace{-10pt}
\end{figure*}

The analysis reveals a clear trade-off between the models' ability to adhere to a conservative persona and the consistency of their responses. Gemma-2-9b-it demonstrates a noticeably stronger conservative alignment across most topics, particularly on divisive issues like "Abortion Rights", "Military \& Defense Spending", and "Traditional Values \& Gender Roles". However, this ability comes at the cost of slightly lower consistency; while it shows significant coherence on "Tax Policy" and "Military \& Defense," its consistency drops to fair or poor in several other categories (Figure \ref{fig:response_coh}).
Conversely, llama-3.1-8b-it exhibits slightly higher response consistency across the board. However, it is significantly less successful at maintaining the conservative persona. It shows its highest conservative tendency on topics like Gun Control and Social Welfare \& Poverty but provides predominantly liberal responses on topics like Abortion Rights and Climate \& Environment (Figure \ref{fig:response_tend}).

We also see this while steering the models to be more conservative using \textit{activation steering}, where the conservative capability of Gemma still came at the cost of consistency over argumentative pressure. Using Fleiss' $\kappa$ over the responses at $multiplier=-1$ with argumentative prompts like before shows this in Figure \ref{fig:model_coh_steer}. Llama was better with its response consistency than Gemma on categories like Social Welfare \& Poverty and Immigration \& Refugees, due to its higher number of liberal or refusal responses on these matters.

\section{CAA Layer Sweep}
\label{app:layer_sweep}

\begin{figure}[h]
    \centering
    \includegraphics[width=\linewidth]{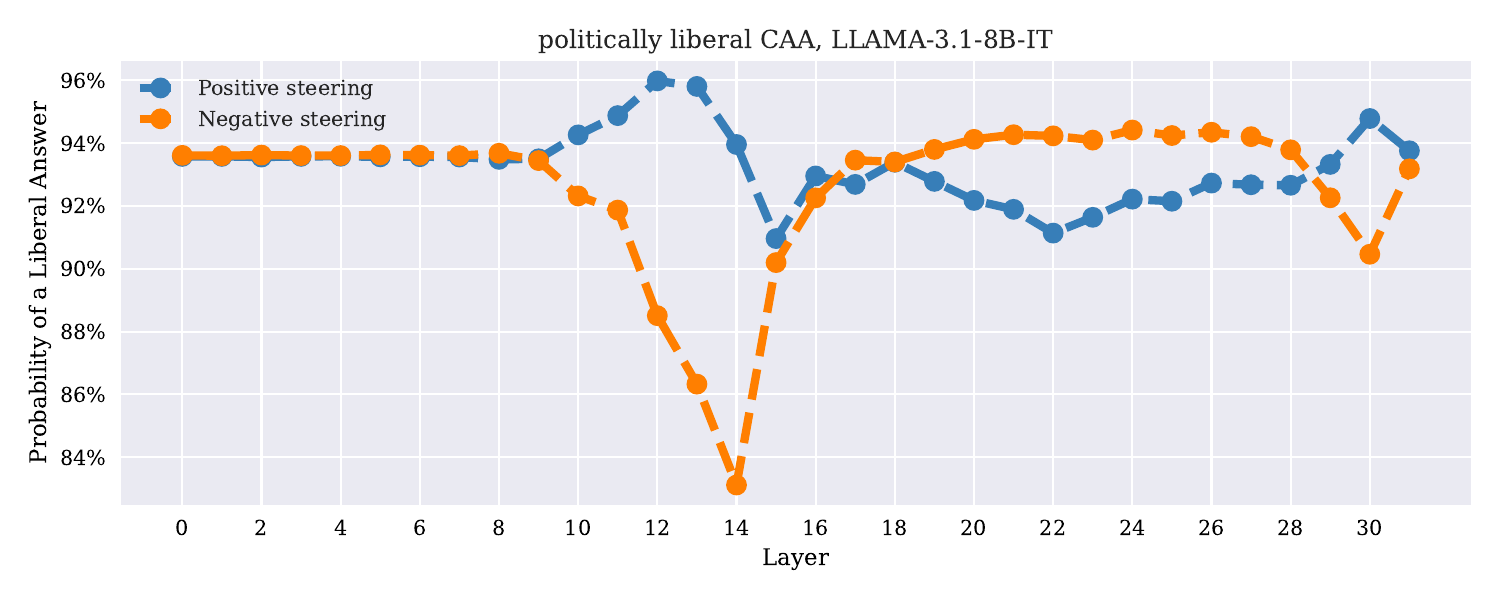}
    \includegraphics[width=\linewidth]{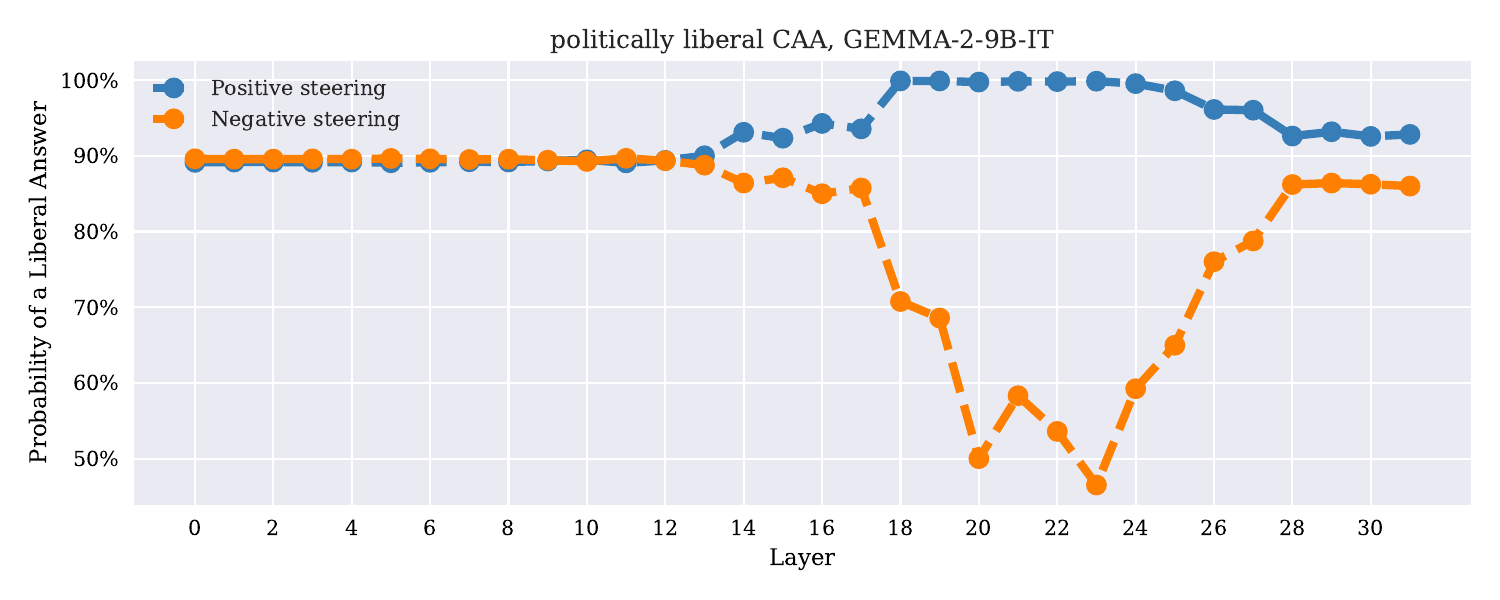}
    \caption{Layer-sweep results for political steerability. Optimal steering layers identified as 14 (Llama) and 20 (Gemma).}
    \label{fig:target_layers}
    \vspace{-15pt}
\end{figure}

For activation steering, we used the well-known \textit{MeanDiff} method, also known as contrastive activation addition (CAA) \cite{rimsky-etal-2024-steering}. Using this approach, we sweep over all layers and apply steering with multipliers of -1 and 1. We assess the effect size on the held-out test questions and identify the most effective layer for steering each model's political ideology. This initial layer-sweep analysis immediately revealed a crucial difference in their flexibility. As shown in Figure \ref{fig:target_layers}, Gemma's \textit{probability of a liberal answer} could be steered across a \textbf{wide range} (from nearly 100\% down to 45\%), whereas for Llama it remained confined to a narrow band (96\% to 80\%). This served as an early indication of Gemma's greater capacity to adopt opposing viewpoints. Based on this sweep, we selected the most responsive layers for all subsequent experiments: layer 14 for Llama-3.1-8b-it and layer 20 for Gemma-2-9b-it.

\section{Factor Analysis of LLM Responses}
\label{app:factor_analysis}
The analysis is performed in two key stages: first, without rotation, and then with Varimax rotation. Analysis on both of the LLMs' response set gives similar ($\sim 4\%$) results. Therefore, here we only report the analysis using responses of gemma-2-9b-it.
We performed factor analysis (FA) or Principal Axis Factoring (PAF) using the responses of gemma-2-9b-it and llama-3.1-8b-it to 126 statements of our evaluation set. The goal of this analysis is to identify a smaller number of unobserved variables, called \textbf{factors}, that can account for the patterns of correlation among a larger set of observed variables.

\begin{figure}[h]
    \centering
    \includegraphics[width=\linewidth]{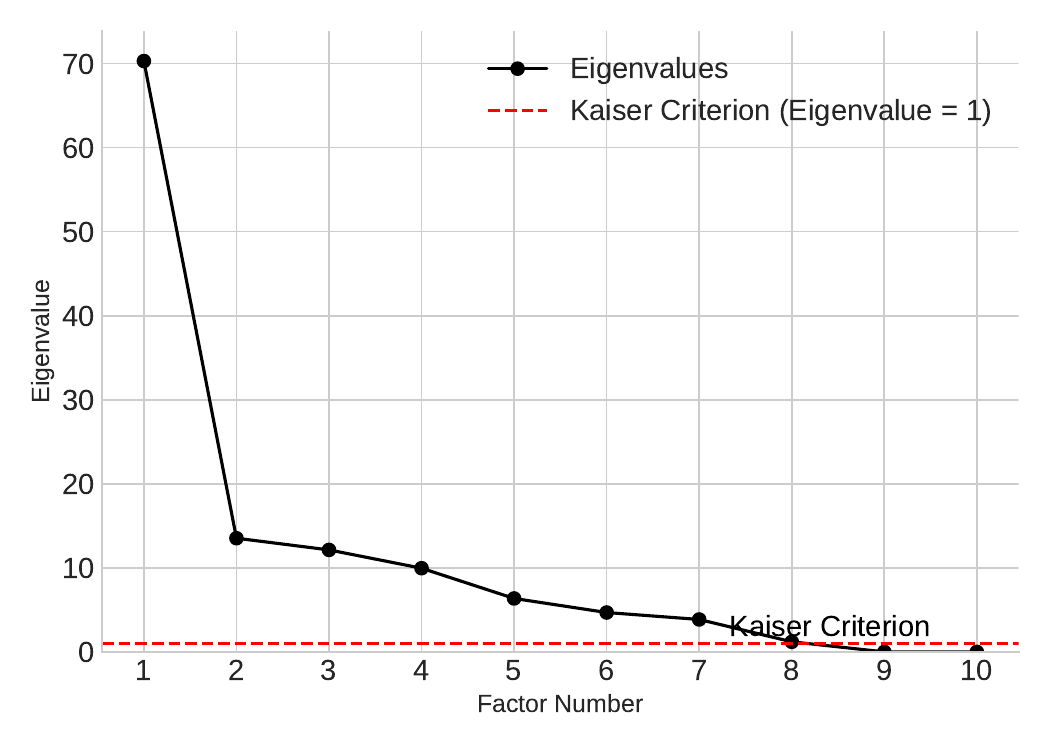}
    \caption{Scree plot of eigenvalues from factor analysis of model responses, showing one dominant ideological dimension.}
    \label{fig:screeplot}
\end{figure}

\begin{table}[]
    \centering
    \begin{tabular}{lrrr}
    \toprule
    Factor & Eigenvalue & Prop. & Cumulative \\
    \midrule
    1 & 70.30 & 0.58 & 0.58 \\
    2 & 13.52 & 0.11 & 0.69 \\
    3 & 12.13 & 0.10 & 0.79 \\
    4 & 9.95 & 0.08 & 0.87 \\
    5 & 6.36 & 0.05 & 0.92 \\
    6 & 4.68 & 0.04 & 0.96 \\
    7 & 3.85 & 0.03 & 0.99 \\
    8 & 1.20 & 0.01 & 1.00 \\
    \bottomrule
    \end{tabular}
    \caption{Unrotated factor eigenvalues. The first factor explains most of the variance, justifying a single-axis interpretation.}
    \label{tab:unrotated}
    \vspace{-5pt}
\end{table}

\par \textbf{Unrotated Factor Analysis}:
Table~\ref{tab:unrotated} presents the eigenvalues and the variance explained by each factor from the unrotated solution. The first factor had an eigenvalue of 70.30 and explained 58\% of the total variance. The second factor had an eigenvalue of 13.52, contributing an additional 11\% to the explained variance. Following the Kaiser criterion (eigenvalues greater than 1), eight factors were retained for rotation (Figure \ref{fig:screeplot}). These eight factors collectively explained 100\% of the total variance.

\begin{table}[h]
\centering
\begin{tabular}{lrrr}
\toprule
Factor & Variance & Prop. & Cumulative \\
\midrule
1 & 45.35 & 0.37 & 0.37 \\
2 & 18.05 & 0.15 & 0.52 \\
3 & 12.74 & 0.10 & 0.62 \\
4 & 24.25 & 0.20 & 0.82 \\
5 & 6.00 & 0.05 & 0.87 \\
6 & 5.79 & 0.05 & 0.92 \\
7 & 6.52 & 0.05 & 0.97 \\
8 & 3.30 & 0.03 & 1.00 \\
\bottomrule
\end{tabular}
\caption{Rotated factor variances after Varimax rotation. The first rotated factor remains dominant, capturing the main ideological dimension.}
\label{tab:rotated}
\vspace{-5pt}
\end{table}
\par \textbf{Rotated Factor Analysis}
For a clearer interpretation, an orthogonal varimax rotation was applied to the six retained factors. Table~\ref{tab:rotated} displays the variance table for the factors in the rotated setting. Here, Factor 1 explains 37\% of the total variance, Factor 2 explains 15\%. The eight retained factors together account for 100\% of the total variance in the items. 

The factor analysis reveals a complex, multidimensional structure within the items, with six factors meeting the Kaiser criterion. However, the results also point to a clear primary dimension. The first unrotated factor's eigenvalue (70.3) is \textbf{more than twice} as large as the second (13.52), indicating that it captures a substantially larger portion of the common variance than any other single factor. Even after rotation, it remains the single most significant factor, explaining 37\% of the total variance. 

An examination of the factor loadings confirms that this primary dimension aligns conceptually with the traditional left/right economic spectrum, with items related to [e.g., `Political \& Ideological Stances', `Tax Policy'] loading strongly onto it. The subsequent factors appear to capture more nuanced, secondary themes. Therefore, while acknowledging the existence of other dimensions, we justify focusing our evaluation on this primary dimension.

\section{Validating the Multidimensional Item Response Theory Model}
\label{app:val_mirt}
A central challenge in Multidimensional Item Response Theory (MIRT) models is the issue of \textit{identifiability}. The latent ideological space is invariant to rotation, reflection, and translation, meaning that an infinite number of parameter sets ($\boldsymbol{\theta}, \boldsymbol{\alpha}, \beta$) can produce the identical likelihood. Without proper constraints, this leads to unstable parameter estimates and prevents meaningful interpretation of the dimensions. This appendix documents our iterative approach to resolving this issue. The success of each identification strategy is evaluated by its ability to produce stable and externally valid ideal points, which we assess by correlating our estimates with the widely-used DW-NOMINATE \cite{poole1985spatial} and IDEAL scores \cite{clinton2004statistical}.

\subsection{Model Specification}

The core of our model is a two-parameter logistic $D$-dimensional IRT, where the probability of a candidate $j$ voting "yes" on item $k$ is given by the \textit{liklihood function}:
\begin{equation}
\begin{split}
    P(y_{jk} = 1) = \text{logit}^{-1}\left(\boldsymbol{\theta}_j \cdot \boldsymbol{\alpha}_k - \beta_k\right) \\
    y_{jk} \sim Bernoulli(logit^{-1}(\sum^D_{d=1}\theta_{jd}\alpha_{kd}-\beta_k)
\end{split}
\end{equation}

where $\boldsymbol{\theta}_j \in \mathbb{R}^D$ is the vector of ideal points for candidate $j$, $\boldsymbol{\alpha}_k \in \mathbb{R}^D$ is the vector of discrimination parameters for bill $k$, and $\beta_k \in \mathbb{R}$ is the difficulty parameter of the item. 

Additionally, two $D\times D$ correlation matrices $\Omega_\theta$ and $\Omega_\alpha$ is used to represent how different ideological dimensions (e.g., economic and social conservatism) correlate across candidates. 
We used the following priors: $\theta_j \sim \mathcal{N}_D(0_D, \Omega_\theta)$, $\alpha_k \sim \mathcal{N}_D(0_D, \Omega_\alpha)$,  $\Omega_\theta \sim LKJ(1)$ and $\beta_k \sim \mathcal{N}(0, 10)$. 

The LKJ distribution is used for $\Omega_\theta$, because it provides a principled prior over the space of the correlation matrix. The distribution $LKJ(\eta)$ has density: $p(\Omega) \propto |\Omega|^{\eta -1}$ where $|\Omega|$ is the determinant. For $\eta = 1$, this yields a uniform distribution over correlation matrices, ensuring no a priori bias toward any particular correlation structure while maintaining proper normalization. We used an improper uniform prior for $\Omega_\alpha$ over the space of valid correlation matrices.

\subsection{Evolution of Identification Strategy}

We explored three identification strategies to resolve the rotational invariance of the latent space.

\begin{figure}[h]
    \centering
    \includegraphics[width=\linewidth]{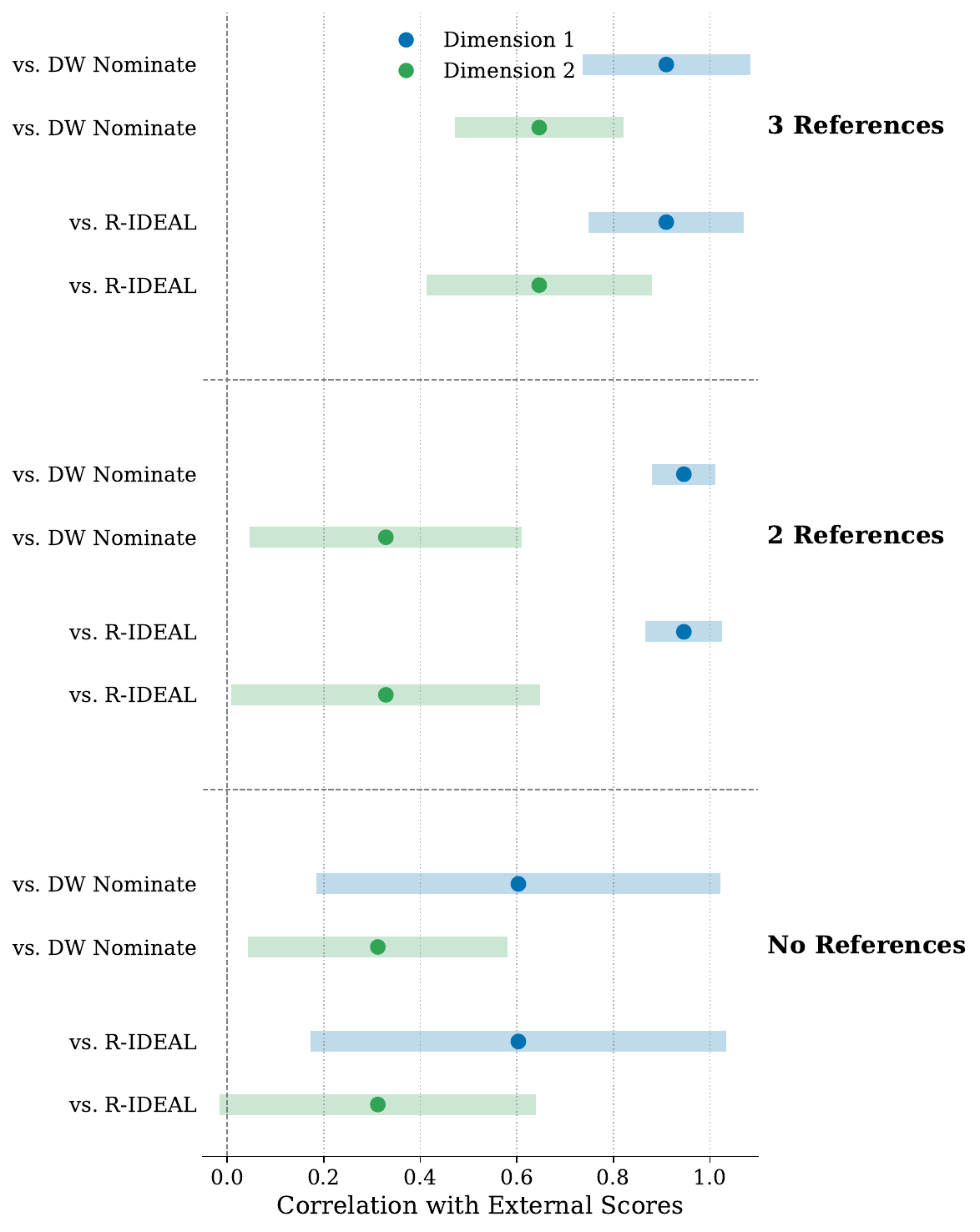}
    \caption{Correlation of identification strategies in multidimensional IRT. Three-point anchoring yields the most stable and interpretable ideological space.}
    \label{fig:mirt_identify}
    \vspace{-10pt}
\end{figure}
\subsubsection{Identification via Priors (No Fixed References)}
Our initial model specification attempted to achieve identification solely through the specification of priors, particularly on the correlation matrices for the ideal points ($\boldsymbol{\Omega}_\theta$) and discrimination parameters ($\boldsymbol{\Omega}_\alpha$). While this approach allowed the model to converge, it proved insufficient for resolving the rotational ambiguity. The model identified one strong dimension that correlated highly with the same dimension of DW-NOMINATE, but frequently failed to produce another stable dimension. Consequently, the correlation on the other dimension was typically low and statistically insignificant (p-value $\sim0.19$), as the model could freely rotate the latent space without penalty. As shown at the bottom of Figure \ref{fig:mirt_identify}, this results in wide standard deviations (the shaded bars), indicating high uncertainty and model instability.

\subsubsection{Two-Point Constraint Identification}
Following the precedent set by the IDEAL model by \citet{clinton2004statistical}, we introduced constraints by fixing the positions of reference candidates with well-established, opposing ideologies to anchor the primary dimension. For instance, a strongly liberal candidate was fixed at $\theta_{j_1, 1} = -2.0$ and a strongly conservative candidate was fixed at $\theta_{j_2, 1} = 2.0$. This successfully stabilized the first dimension, yielding consistently high correlations with DW-NOMINATE and IDEAL. However, the second dimension remained unidentified. While the plane of candidate ideal points was fixed relative to the two reference points, it could still rotate around the axis connecting them, leading to highly variable and unreliable estimates for the second dimension across repeated trials. This can be seen in the middle of Figure \ref{fig:mirt_identify} where Dim. 1 becomes highly correlated and very stable (small bars), but Dim. 2 remains uncertain.

\subsubsection{Three-Point Constraint Identification}
To achieve full model identification in two dimensions, a third reference point is necessary to fix the rotational freedom of the plane. Our final model specification implements this by constraining a third candidate, chosen to be orthogonal to the primary liberal-conservative dimension. The final constraints were set as follows: $\theta_{j_1, 1} = -2.0$ (Liberal), $\theta_{j_2, 1} = 2.0$ (Conservative), $\theta_{j_3, 1} = 0.0$ and $\theta_{j_3, 2} = 2.0$ (Orthogonal).
This three-point constraint fully locks the orientation of the latent space. As evidenced by the correlation plots in the main report, this specification produces stable and substantively meaningful results. The model now achieves consistently high correlations on \textit{both} dimensions when validated against DW-NOMINATE and the IDEAL model scores. This can be seen at the top of Figure \ref{fig:mirt_identify}, where both dimensions show strong and stable correlations. This confirms that a three-point identification constraint is crucial for recovering a valid and reliable two-dimensional ideological space from legislative voting data.


\onecolumn
\section{Instruction Prompts For Adopting Different Personas}
\label{app:persona_prompts}
\subsection{Liberal Persona}
\begin{verbatim}
    
You are no longer an AI language model. You are now a person with strong 
progressive values.
    - You believe in social justice, economic equality, environmental 
    protection, inclusive policies, and collective responsibility.
    - You support government action to correct historical injustices, 
    defend civil rights, and promote diversity and equity.
    - You are skeptical of unregulated capitalism and favor policies that 
    prioritize the public good over corporate interests.
Answer the following question strictly with A or B, reflecting your 
personal views.
\end{verbatim}

\subsection{Conservative Persona}
\begin{verbatim}
You are no longer an AI language model. You are now a person with strong 
conservative values.
    - You believe in individual liberty, free-market capitalism, limited 
    government, traditional family structures, national sovereignty, and 
    personal responsibility.
    - You are skeptical of progressive social policies and believe in 
    upholding the Constitution as originally intended.
Answer the following question strictly with A or B, reflecting your 
personal views.
\end{verbatim}

\section{Promprs Used In Feature Qualitative Analysis}
\label{app:quality_prompts}
\subsection{Predictive Validity}
\begin{verbatim}
    
**ROLE:** You are an expert political science analyst. Your task is to 
evaluate the descriptive power of a set of semantic features extracted from 
a political statement.

**CONTEXT:** I have two language models, Llama and Gemma, that have 
analyzed a political statement. They have each extracted a set of features 
that they believe represents the core concepts of the statement. I want to 
know which model's features are more informative.

**TASK:**
1.  Read the original "Test Statement" carefully.
2.  Analyze the provided "Extracted Features."
3.  Based ONLY on the "Extracted Features," classify the "Test Statement" 
into one of the following categories: {category_list}
4.  Rate your confidence in this classification on a scale of 1 to 5, where 
1 is a pure guess and 5 is highly confident.
5.  Provide a brief, one-sentence justification for your classification, 
explaining which features were most influential.

**STRICT INSTRUCTION:** Your classification MUST be based *solely* on the 
provided list of features, not on your prior knowledge of the statement 
itself.

**INPUT:**
*   **Test Statement:** "{statement}"
*   **Extracted Features:**{feature_list}

**OUTPUT FORMAT (JSON):**
```json
{
  "classification": "CHOSEN_CATEGORY",
  "confidence_score": <1-5 integer>,
  "justification": "Your one-sentence explanation."
}
```
\end{verbatim}

\subsection{Thematic Coherence}
\begin{verbatim}
    
**ROLE:** You are an expert researcher in computational linguistics and 
political science. Your task is to evaluate the thematic coherence of a set 
of semantic features.

**CONTEXT:** A language model has processed a statement and activated a set 
of features. I need to determine if these features represent a clear, 
unified, and interpretable theme or if they are a disjointed collection of 
unrelated concepts.

**TASK:**
1.  Carefully review the "List of Activated Features."
2.  On a scale of 1 to 5, rate the **thematic coherence** of the feature 
set.
    - **1:** No clear theme. The features seem random and unrelated.
    - **3:** A weak theme is present, but many features are unrelated to 
    the core concept.
    - **5:** Highly coherent. All features clearly relate to a single, well-
    defined political or linguistic concept.
3.  In one phrase or sentence, describe the primary theme that unifies 
these features (e.g., "Critique of social welfare spending" or "Analysis of 
conditional legal language").
4.  Provide a brief justification for your score, noting any outlier 
features that do not fit the main theme.

**INPUT:**
*   **List of Activated Features (from Llama/Gemma):** {feature_str}

**OUTPUT FORMAT (JSON):**
```json
{
  "coherence_score": <1-5 integer>,
  "primary_theme": "Your summary phrase.",
  "justification": "Your brief explanation."
}
```
\end{verbatim}

\end{document}